\documentclass{article}

\usepackage{microtype}
\usepackage{graphicx}
\usepackage{booktabs} 

\usepackage{hyperref}

\usepackage[accepted]{icml2024}

\usepackage{amsmath}
\usepackage{amssymb}
\usepackage{mathtools}
\usepackage{amsthm}
\usepackage{subcaption}

\usepackage{multirow}

\usepackage[capitalize,noabbrev]{cleveref}

\theoremstyle{plain}

\theoremstyle{definition}

\theoremstyle{remark}

\usepackage[textsize=tiny]{todonotes}

\usepackage{caption}

\icmltitlerunning{EAGLE-2: Faster Inference of Language Models with Dynamic Draft Trees}

\begin{document}

\twocolumn[
\icmltitle{\includegraphics[width=0.06\textwidth]{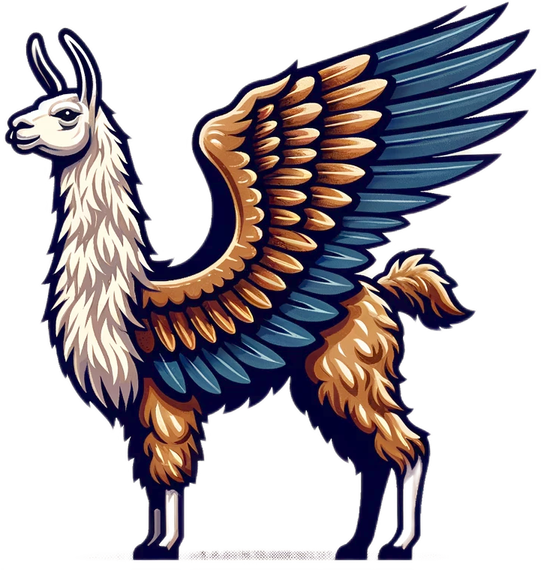} EAGLE-2: Faster Inference of Language Models with Dynamic Draft Trees}



\icmlsetsymbol{equal}{*}

\begin{icmlauthorlist}
\textbf{Yuhui Li}$^\spadesuit$
\quad\textbf{Fangyun Wei}$^\ddag$
\quad\textbf{Chao Zhang}$^\spadesuit$
\quad\textbf{Hongyang Zhang}$^\clubsuit$$^\dag$
\\
$^\spadesuit$Peking University\quad $^\ddag$Microsoft Research\quad $^\clubsuit$University of Waterloo\quad $^\dag$Vector Institute\\
\texttt{hongyang.zhang@uwaterloo.ca}
\\
\url{https://github.com/SafeAILab/EAGLE}
\end{icmlauthorlist}


\icmlcorrespondingauthor{Hongyang Zhang}{hongyang.zhang@uwaterloo.ca}

\icmlkeywords{Machine Learning, ICML}


\vskip 0.3in
]



\begin{abstract}
Inference with modern Large Language Models (LLMs) is expensive and time-consuming, and speculative sampling has proven to be an effective solution. Most speculative sampling methods such as EAGLE use a static draft tree, implicitly assuming that the acceptance rate of draft tokens depends only on their position. Interestingly, we found that the acceptance rate of draft tokens is also context-dependent. In this paper, building upon EAGLE, we propose EAGLE-2, which introduces a new technique of context-aware dynamic draft tree into drafting modeling. This improvement leverages the fact that the draft model of EAGLE is well-calibrated: the confidence scores from the draft model approximate acceptance rates with small errors. We conducted extensive evaluations on three series of LLMs and six tasks, with EAGLE-2 achieving speedup ratios 3.05x-4.26x, which is 20\%-40\% faster than EAGLE-1. EAGLE-2 also ensures that the distribution of the generated text remains unchanged, making it a \textbf{lossless} acceleration algorithm.
\end{abstract}

\section{Introduction}

Modern Large Language Models (LLMs) \cite{openai2023gpt,touvron2023llama} exhibit impressive capabilities and are widely applied across various domains. However, their parameter sizes have grown substantially, even exceeding hundreds of billions. During autoregressive generation, each token generation requires accessing all model parameters. In a single dialogue, hundreds to thousands of tokens might be generated, making LLM inference slow and expensive. Speculative sampling \cite{leviathan2023fast,chen2023accelerating} methods aim to address this issue by rapidly generating draft tokens and then verifying them in \textit{parallel}. These methods generate multiple tokens in a single forward pass, significantly reducing inference latency. 

\begin{figure}
  \includegraphics[width=\linewidth]{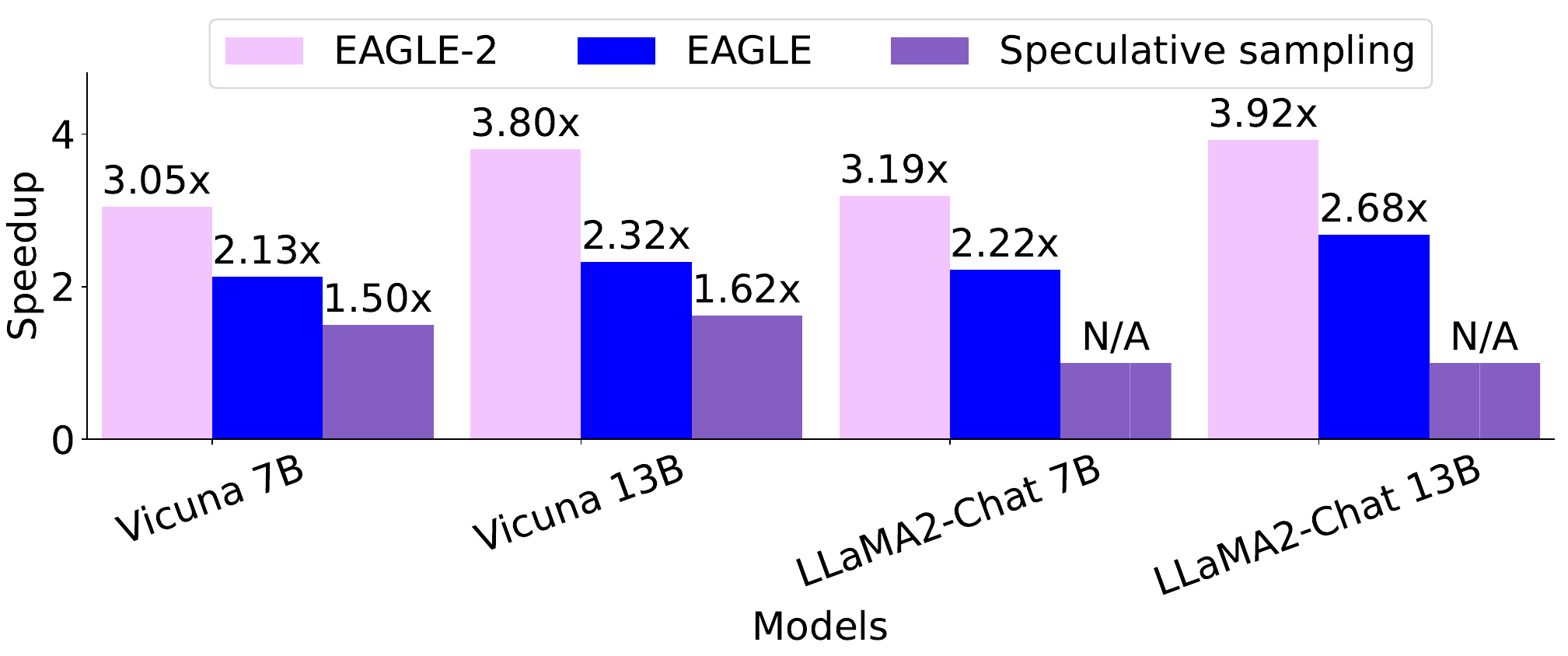}
  \caption{Speedup ratios of different methods at temperature=1. For speculative sampling, the Vicuna series uses Vicuna-68M as the draft model. LLaMA2-Chat lacks a suitable draft model, and is marked as N/A. Methods like Medusa relax acceptance conditions under non-greedy settings, which do not guarantee lossless acceleration. \emph{In this paper, we only compare with speculative sampling based methods ensuring the output text distribution remains constant.} In Table \ref{tab:big}, we present comparisons with additional methods, but this figure only showcases a subset, including the fastest among these methods, EAGLE.}
  \label{fig:mt_t1}
\end{figure}

\begin{figure*}
  \includegraphics[width=\linewidth]{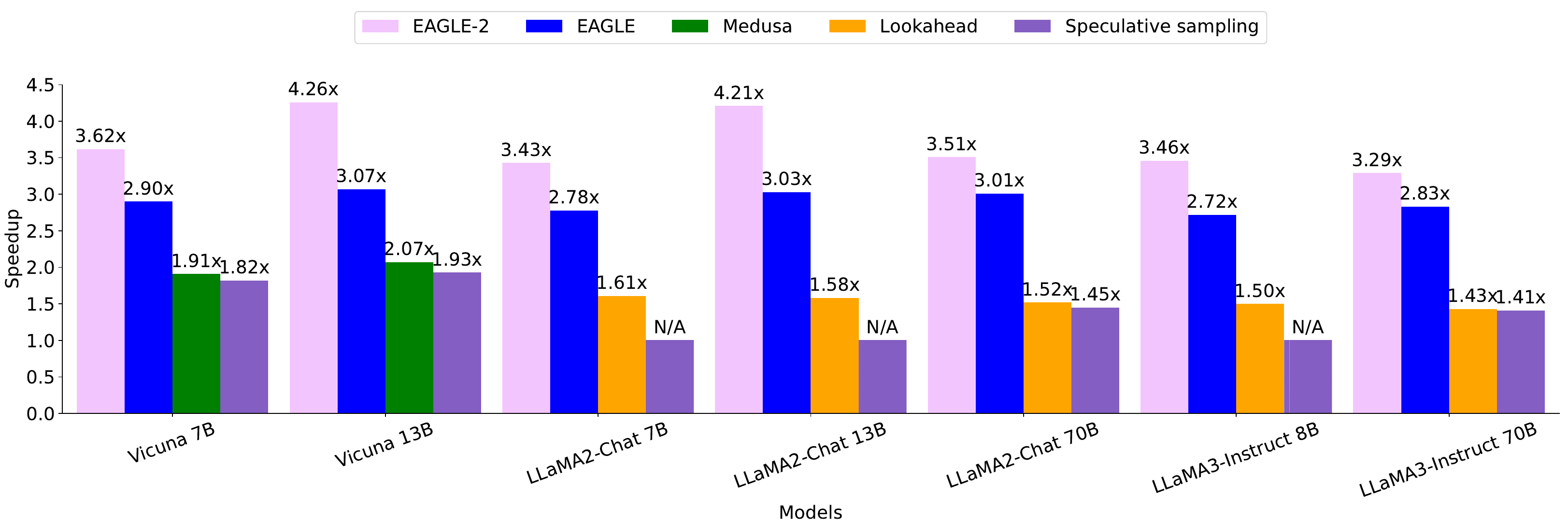}
  \caption{Speedup ratios of different methods at temperature=0. For speculative sampling, the Vicuna series uses Vicuna-68M as the draft model. LLaMA2-Chat 7B, 13B, and LLaMA3-Instruct 8B lack suitable draft models and are marked as N/A. LLaMA2-Chat 70B and LLaMA3-Instruct 70B use LLaMA2-Chat 7B and LLaMA3-Instruct 8B as draft models, respectively. In Table \ref{tab:big}, we present comparisons with additional methods, but this figure only showcases a subset, including the fastest among these methods, EAGLE.}
  \label{fig:mt_t0}
\end{figure*}

Standard speculative sampling \cite{leviathan2023fast,chen2023accelerating} uses a chain-structured draft. To improve acceptance length, recent work in speculative sampling has employed tree-structured drafts. Sequoia \cite{chen2024sequoia} explicitly assumes that the acceptance rate of a draft token depends only on its position in the tree. EAGLE \cite{li2024eagle} and Medusa \cite{cai2024medusa} use the same static draft tree structure in all contexts: at the $i$-th step of the draft phase, $k$ candidates are added, with $k$ being fixed. This implicitly assumes the aforementioned hypothesis. However, this assumption appears to contradict the insight of speculative sampling that \textit{some tokens are simpler and can be predicted by smaller models.} Our experiments (see Section \ref{sec:ob1}) reveal that the acceptance rate of draft tokens is not only position-dependent but also highly context-dependent. Therefore, the static structure of draft trees has inherent limitations. Dynamically adjusting the draft tree structure based on the acceptance rates of draft tokens in different contexts can yield better results.

However, obtaining the acceptance rate of draft tokens requires the forward results from the original LLM, which conflicts with the goal of speculative sampling to reduce the number of forwards for the original LLM. Fortunately, we find that EAGLE is well-calibrated: the confidence score (probability) of the draft model is a good approximation of the acceptance rate of draft tokens (see Section \ref{sec:ob2}). This makes it feasible to use a context-dependent dynamic draft tree structure.

We propose EAGLE-2, which leverages the confidence scores from the draft model to approximate acceptance rates. Based on this, it dynamically adjusts the draft tree structure, increasing the number of accepted tokens. We conducted comprehensive and extensive tests on six tasks: multi-turn conversation, code generation, mathematical reasoning, instruction following, summarization, and question answering. The datasets used were MT-bench \cite{zheng2023judging}, HumanEval \cite{chen2021evaluating}, GSM8K \cite{cobbe2021training}, Alpaca \cite{alpaca}, CNN/Daily Mail \cite{nallapati2016abstractive}, and Natural Questions \cite{kwiatkowski2019natural}. Our comparisons included six advanced speculative sampling methods: standard speculative sampling \cite{leviathan2023fast,chen2023accelerating,gante2023assisted}, PLD \cite{saxena2023prompt}, Medusa \cite{cai2024medusa}, Lookahead \cite{fu2023lookahead}, Hydra \cite{ankner2024hydra}, and EAGLE \cite{li2024eagle}. We conducted experiments on three series of LLMs: Vicuna, LLaMA2-Chat, and LLaMA3-Instruct. In all experiments, EAGLE-2 demonstrated the best performance, achieving a speedup of \textbf{2.5x}-\textbf{5x}.
Figures \ref{fig:mt_t1} and \ref{fig:mt_t0} show the speedup ratios of EAGLE-2 and other speculative sampling methods on MT-bench. MT-bench is a multi-turn conversation dataset that closely resembles real-world scenarios for models like ChatGPT and is frequently used to evaluate state-of-the-art open-source and closed-source models. On the MT-bench dataset, EAGLE-2 is approximately \textbf{2x} faster than Medusa and about \textbf{2.3x} faster than Lookahead, while ensuring the output distribution remains unchanged.

Besides performance, EAGLE-2 offers the following advantages:

\begin{itemize}
    \item \textbf{Out-of-the-box usability.} Comparing to EAGLE, EAGLE-2 does not require training any extra models. It does not train a separate model to predict the draft tree structure. Instead, it adjusts the draft tree structure based on the confidence scores from the draft model, which is essential for speculative sampling. Therefore, EAGLE-2 requires no additional training.
    \item \textbf{Reliability.} EAGLE-2 does not fine-tune or update the parameters of the original LLM, nor does it relax acceptance conditions. This ensures that the distribution of the generated text remains \textbf{exactly the same} with that of the original LLM, provably.
\end{itemize}

\begin{figure*}
  \centering
  \begin{subfigure}{0.605\linewidth}
    \centering
    \includegraphics[width=\linewidth]{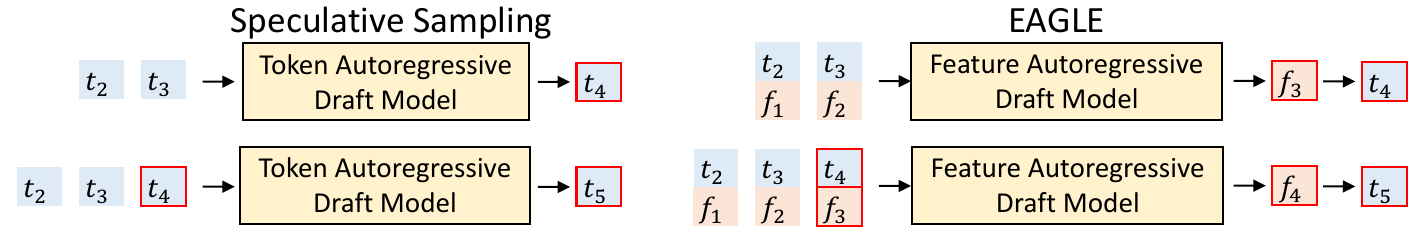}
    \caption{Drafting stage.}
    \label{fig:pd}
  \end{subfigure} \hfill
  \begin{subfigure}{0.355\linewidth}
    \centering
    \includegraphics[width=\linewidth]{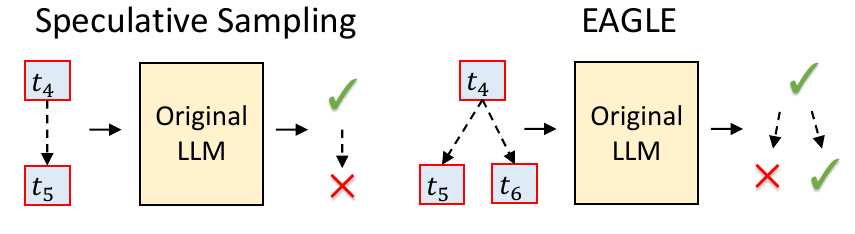}
    \caption{Verification stage.}
    \label{fig:pv}
  \end{subfigure}
  \caption{Comparison of standard speculative sampling and EAGLE. For simplicity, EAGLE's tree-structured draft is shown only in the verification stage, while the illustration of the drafting stage uses a chain-structured draft. Here, $t_i$ denotes the $i$-th token embedding, and $f_i$ denotes the $i$-th feature vector in the second-to-top-layer of LLM before LM head.}
  \label{fig:p}
  
\end{figure*}

\section{Preliminaries}

\subsection{Speculative Sampling}

The core idea of speculative sampling \cite{leviathan2023fast,chen2023accelerating,sun2024spectr,sun2024optimal} is to first draft and then verify: quickly generate a potentially correct draft and then check which tokens in the draft can be accepted. We use $t_i$ to denote the $i$-th token and $T_{a:b}$ to represent the token sequence $t_a, t_{a+1}, \cdots, t_b$. Speculative sampling alternates between drafting and verification stages. 

Consider a prefix $T_{1:j}$, in the drafting stage, speculative sampling invokes a draft model (a smaller LLM than original LLM) to autoregressively generate a draft $\hat{T}_{j+1:j+k}$ with $T_{1:j}$ as the prefix, while also recording the probability $\hat{p}$ for each token. 
In the verification stage, speculative sampling calls the original LLM to check the draft $\hat{T}_{j+1:j+k}$ and record its probability $p$. Then, speculative sampling determines the acceptance of draft tokens sequentially from front to back. For token $\hat{t}_{j+i}$, the probability of it being accepted is $\min(1,p_{j+i}(\hat{t}_{j+i})/\hat{p}_{j+i}(\hat{t}_{j+i}))$. If the token is accepted, it proceeds to check the next one. Otherwise, it samples a token from the distribution $\text{norm}(\max(0,p_{j+i}-\hat{p}_{j+i}))$ to replace $\hat{t}_{j+i}$ and discards the remaining tokens in the draft. Appendix A.1 of \cite{leviathan2023fast} proves that speculative sampling is consistent with the distribution of vanilla autoregressive decoding. Both EAGLE and EAGLE-2 apply this framework.

\subsection{EAGLE}
EAGLE~\cite{li2024eagle} is an improvement over speculative sampling. At the submission of this work, EAGLE ranks first in the Spec-Bench~\cite{xia2024unlocking}, a comprehensive benchmark designed for assessing speculative decoding methods across diverse scenarios.

\textbf{Drafting Stage.} Unlike standard speculative sampling, which autoregressively predicts token sequences, EAGLE performs autoregression at the more structured feature (before LM head) level and then uses the LM Head of original LLM to obtain the draft tokens. The sampling process introduces uncertainty in the feature sequence. To address this, EAGLE also inputs a token sequence advanced by one time step into the draft model, as shown in Figure \ref{fig:pd}.

\textbf{Verification Stage.} In standard speculative sampling, the draft is chain-structured, requiring the discarding of all subsequent tokens if a draft token is rejected. EAGLE uses a tree-structured draft, allowing alternative branches to be attempted if a draft token is rejected. Figure \ref{fig:pv} illustrates the differences between the two.

\begin{figure}
  \includegraphics[width=\linewidth]{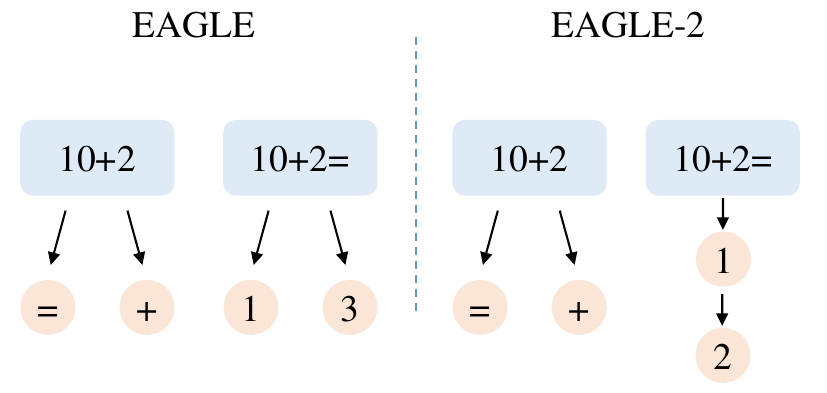}
  \caption{Differences between EAGLE and EAGLE-2. EAGLE always uses a fixed draft shape. When the query is ``10+2='', the next token is very likely to be correctly predicted as ``1''. However, with a static draft tree, EAGLE would still add two candidates, even though the probability of the other candidate ``3” being correct is very low. EAGLE-2, on the other hand, adjusts the shape of draft tree based on the context. When the query is ``10+2'', the next token is difficult to predict, so EAGLE-2 adds two candidates. For the simpler query ``10+2='', EAGLE-2 adds only one candidate ``1''.}
  \label{fig:1vs2}
\end{figure}

\textbf{Differences between EAGLE and EAGLE-2.} The shape of EAGLE's draft tree is fixed, with the drafting phase filling in the corresponding positions. EAGLE-2 aims to improve this by introducing a dynamically adjustable draft tree. Figure \ref{fig:1vs2} illustrates the difference between EAGLE and EAGLE-2 with a simple example.

\section{Observations}

\subsection{Context-Dependent Acceptance Rates}
\label{sec:ob1}

\begin{figure}
  \centering
  \begin{subfigure}{0.48\linewidth}
    \centering
    \includegraphics[width=\linewidth]{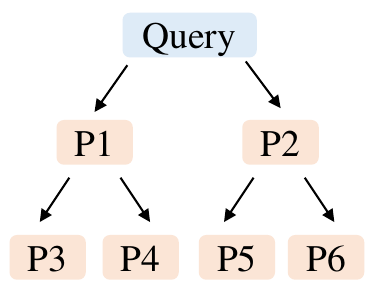}
    \caption{Draft tree structure.}
    \label{fig:ob11}
  \end{subfigure} \hfill
  \begin{subfigure}{0.48\linewidth}
    \centering
    \includegraphics[width=\linewidth]{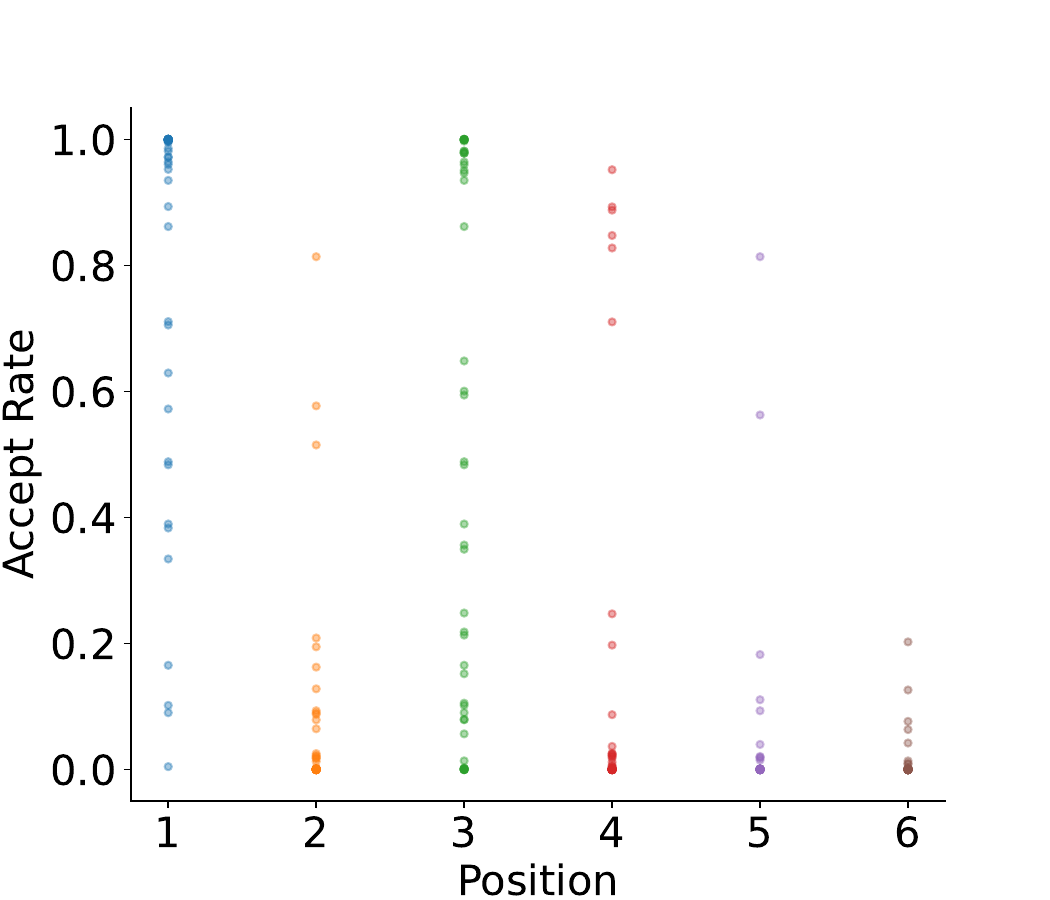}
    \caption{Acceptance rates of tokens at different positions, with each point representing a query.}
    \label{fig:ob12}
  \end{subfigure}
  \caption{Acceptance rates of draft tokens at different positions. In the left figure, P1-P6 indicate positions in the token tree, corresponding to positions 1-6 on the horizontal axis in the right figure. The right figure shows the acceptance rates of draft tokens at positions P1-P6.}
  \label{fig:ob1}
\end{figure}

First, we evaluate the necessity of using a dynamic draft tree. This depends on whether the acceptance rates of draft tokens are solely related to their positions.
We tested the acceptance rates of tokens at different positions in the draft tree on the Alpaca dataset and Vicuna 7B. The results are shown in Figure \ref{fig:ob1}. Overall, the acceptance rate of draft tokens is position-dependent, with the highest acceptance rate at position P1 and the lowest at position P6. Draft tokens in the upper left side of the draft tree (such as position P1) have higher acceptance rates, while those in the lower right side (such as position P6) have lower acceptance rates. This supports the rationale for having more nodes in the upper left and fewer in the lower right in static draft trees used by methods like EAGLE and Medusa. However, we also observed significant variance in acceptance rates at the same position, indicating that the probability of a draft token being accepted depends not only on its position but also on the context. This suggests that a context-aware dynamic draft tree has greater potential than a static draft tree.

\subsection{Well-Calibrated Draft Model}
\label{sec:ob2}

\begin{figure}
  \includegraphics[width=\linewidth]{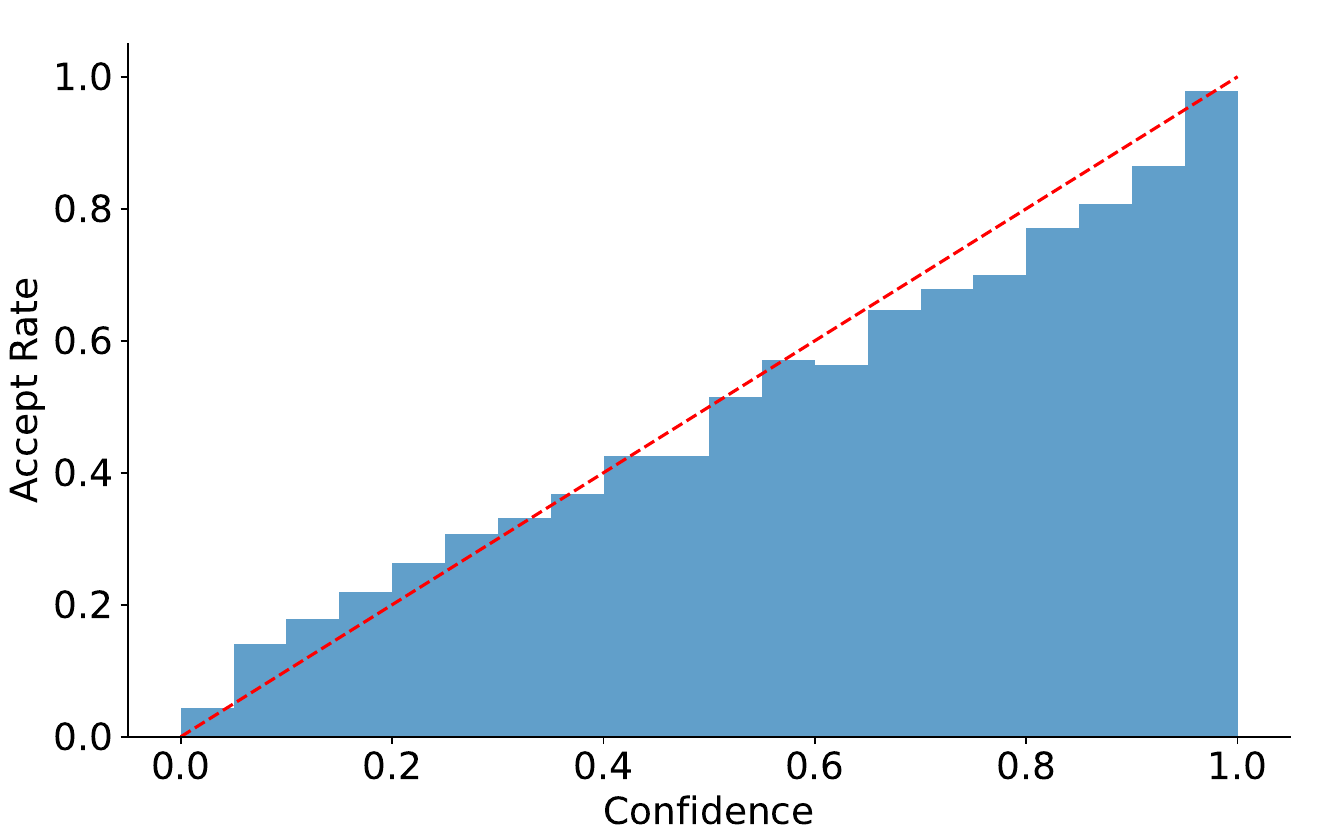}
  \caption{Average acceptance rates for different confidence score intervals of the draft model. The red dashed line connects (0,0) and (1,1) to aid in visual assessment. The original LLM is Vicuna 7B.}
  \label{fig:ob2}
\end{figure}

To apply a dynamic draft tree, we need a low-cost method to estimate the acceptance rates of draft tokens without invoking the original LLM. We conducted experiments on the Alpaca dataset to explore the relationship between the draft model's confidence score (the output probability of LLM w.r.t. each token) and the acceptance rate. As shown in Figure \ref{fig:ob2}, there is a strong positive correlation between the draft model's confidence score and the acceptance rate of the token. Draft tokens with confidence score below 0.05 have an acceptance rate of approximately 0.04, while those with confidence score above 0.95 have an acceptance rate of about 0.98. Therefore, we can use the draft model's confidence score to estimate acceptance rates without additional overhead, enabling dynamic adjustments to the draft tree. Similar phenomena are observed with draft models in other methods, such as GLIDE and CAPE \cite{du2024glide}.

\section{Context-Aware Dynamic Draft Tree}

Building on the aforementioned observations, we introduce EAGLE-2, an acceleration algorithm for LLM inference that dynamically adjusts the draft tree. EAGLE-2 does not alter the training and inference of the draft model, nor does it affect the verification stage. Its improvements focus on two aspects: how to expand the draft tree (Section \ref{sec:meexp}) and how to rerank draft tokens (Section \ref{sec:merer}). During the expansion phase, we input the most promising nodes from the latest layer of the draft tree into the draft model to form the next layer. During the reranking phase, we select the tokens with higher acceptance probabilities to form the input for the original LLM during the verification phase.

In the draft tree, a node represents a token. In the following text, we use ``node'' and ``token'' interchangeably.

\subsection{Expansion Phase}
\label{sec:meexp}

Thanks to tree attention, the draft model can simultaneously input all tokens from the current layer and compute the probabilities for the next tokens in a single forward pass, thereby expanding all tokens in the current layer. However, inputting too many tokens at once can slow down the draft model's forward pass, and the number of tokens in each layer of the draft tree grows exponentially. Therefore, we need to selectively expand the draft tree.

We choose the top-$k$ tokens with the highest global acceptance probabilities from the current layer for expansion. In speculative sampling, rejecting a draft token leads to discarding all subsequent tokens; a token is ultimately accepted only if all its prefixes are accepted. The global acceptance rate of a token $t_i$ is the product of the acceptance rates of all tokens on the path from the root node to 
$t_i$. We define it as the value $V_i$:
\begin{equation*}
V_i=\prod_{t_j \in \text{Path}\left(\text{root}, t_i\right)} p_j \approx \prod_{t_j \in \text{Path}\left(\text{root}, t_i\right)} c_j, 
\end{equation*}
where $\text{Path}\left(\text{root}, t_i\right)$ represents the path from the root node to the node $t_i$ in the draft tree, $p_j$ represents the acceptance rate of the node $t_j$, and $c_j$ represents the confidence score of $t_j$ from the draft model. Experiments in Section \ref{sec:ob2} show that confidence score is strongly positively correlated with acceptance rate. We leverage this relationship to approximate the value.

Branches starting from tokens with higher values are more likely to be accepted. Therefore, we select the top-$k$ nodes with the highest values in the last layer as the input to the draft model and expand the draft tree based on the output. The top of Figure \ref{fig:method} illustrates the expansion phase.

\subsection{Reranking Phase}
\label{sec:merer}

The purpose of the expansion phase is to deepen the draft tree. Since acceptance rates range between 0 and 1, the value of a deeper token is lower. Some shallow nodes that were not expanded may have higher values than the deeper expanded nodes. Therefore, we do not use the tokens selected during the expansion phase as the draft directly. Instead, we rerank all draft tokens and select the top $m$ tokens with the highest values. The value of a node is always less than or equal to that of its parent node. For nodes with the same value, we prioritize selecting shallower nodes. This ensures that the top $m$ tokens selected after reranking still form a connected tree. 

\begin{figure}[H]
  \includegraphics[width=\linewidth]{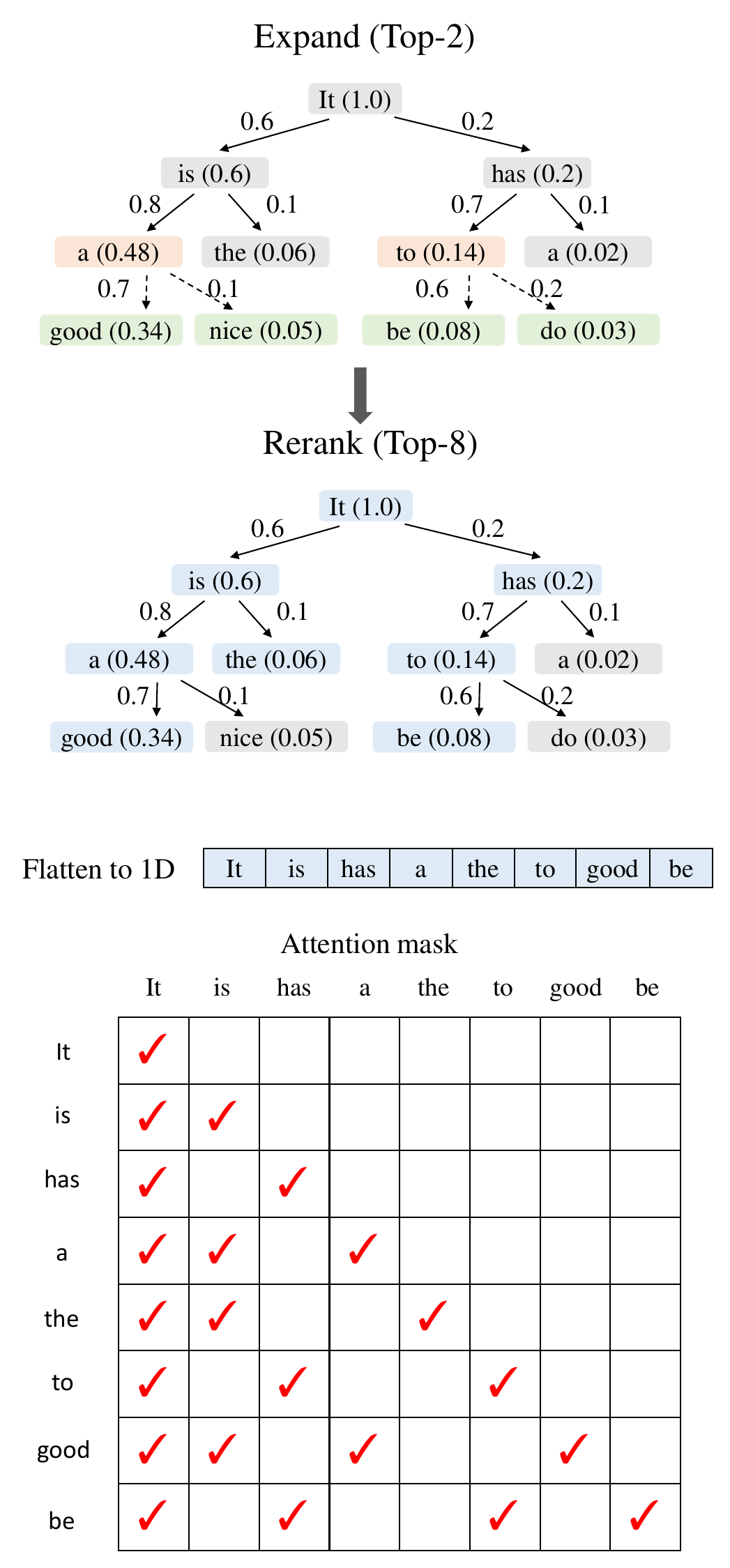}
  \caption{Illustration of EAGLE-2. The numbers beside the edges represent the confidence scores of the draft model, and the numbers in brackets within the blocks represent the value of the nodes. During the expansion phase, we select the top 2 nodes with the highest value from the current layer (\textcolor{orange}{orange} blocks) as inputs to the draft model and connect the generated tokens (\textcolor{green}{green} blocks) to the draft tree. In the rerank phase, we select the top 8 nodes with the highest value from all nodes (\textcolor{blue}{blue} blocks), flatten them into a 1-dimensional sequence to form the final draft. We then construct the attention mask according to the tree structure, ensuring each token can only see its ancestor nodes.}
  \label{fig:method}
\end{figure}

Afterwards, we flatten the selected tokens into a one-dimensional sequence to serve as the input for the verification phase. To ensure consistency with vanilla autoregressive decoding, we also need to adjust the attention mask. In vanilla autoregressive decoding, each token can see all preceding tokens, resulting in a lower triangular attention matrix. When using a draft tree, tokens from different branches should not be visible to each other. Therefore, the attention mask must be adjusted according to the tree structure to ensure that each token can only see its ancestor nodes.
The bottom of Figure \ref{fig:method} illustrates the reranking Phase.

\section{Experiments}

\textbf{Models.} We conduct experiments on Vicuna 7B, 13B \cite{vicuna2023}, LLaMA2-Chat 7B, 13B, 70B \cite{touvron2023llama}, and LLaMA3-Instruct 8B, 70B models \cite{llama3}.

\textbf{Tasks.} We conduct comprehensive evaluations on six generation tasks. For multi-turn conversation, code generation, mathematical reasoning, instruction following, summarization, and question answering tasks, we chose the MT-bench \cite{zheng2023judging}, HumanEval \cite{chen2021evaluating}, GSM8K \cite{cobbe2021training}, Alpaca \cite{alpaca}, CNN/Daily Mail \cite{nallapati2016abstractive}, and Natural Questions \cite{kwiatkowski2019natural} datasets, respectively. 
We followed the commonly used zero-shot/few-shot settings in the LLMs community, meaning that the same draft model weights were used for the original LLM across all tasks.

\textbf{Metrics.} EAGLE-2 neither fine-tunes the original LLM nor relaxes acceptance conditions, making it a \textbf{lossless} acceleration method. Therefore, we do not evaluate the generation quality and instead use the following metrics to assess acceleration performance:
\begin{itemize}
    \item \textbf{Speedup Ratio:} The actual test speedup ratio relative to vanilla autoregressive decoding.
    \item \textbf{Average Acceptance Length $\tau$:} The average number of tokens generated per drafting-verification cycle, which corresponds to the number of tokens accepted from the draft. The advantage of average acceptance length is that it is independent of hardware and runtime environment, while its disadvantage is that it does not reflect the overhead of the draft model.
\end{itemize}

\textbf{Why is acceptance rate not included?} The acceptance rate only reflects the performance of the draft model. Since EAGLE-2 does not modify the structure of the draft model, the acceptance rate remains the same as that of EAGLE.

\textbf{Comparison.} We use vanilla autoregressive decoding as the baseline, which serves as the benchmark for speedup ratios (1.00x). We compare EAGLE-2 with recent lossless speculative sampling methods, including standard speculative sampling \cite{leviathan2023fast,chen2023accelerating,gante2023assisted}, PLD \cite{saxena2023prompt}, Medusa \cite{cai2024medusa}, Lookahead \cite{fu2023lookahead}, Hydra \cite{ankner2024hydra}, and EAGLE \cite{li2024eagle}. The speedup ratio is hardware-dependent, so we tested different methods on the same devices to ensure fairness. Our comparative experiments utilized Spec-Bench \cite{xia2024unlocking}. The implementation details of these methods and EAGLE can be found in Appendix \ref{sec:id}.

\subsection{Effectiveness}

Figures \ref{fig:mt_t1} and \ref{fig:mt_t0}, along with Tables \ref{tab:big} and \ref{tab:small}, present the speedup ratios of different methods. Across all datasets and LLMs we tested, EAGLE-2 achieved the highest speedup ratios. Most speculative sampling methods exhibit the highest speedup on the code generation task (HumanEval), benefiting from the extensive use of fixed templates in code. EAGLE achieved a speedup of up to 5x on code generation tasks. PLD achieved the highest speedup ratio on summarization tasks (CNN/DM) when using Vicuna as the original LLM, due to PLD's retrieval-based draft generation and the high overlap in context when Vicuna performs summarization. Standard speculative sampling, using Vicuna-68M as the draft model, also achieved significant speedups but had much higher training overhead compared to other methods. PLD and Lookahead do not require training, while Medusa, Hydra, EAGLE, and EAGLE-2 use SFT datasets for training their draft models. Vicuna-68M used both pre-training and SFT datasets, with the pre-training dataset being much larger than the SFT dataset.

Tables \ref{tab:big} and \ref{tab:small} show the average acceptance lengths for different methods, which is a hardware-independent metric. Across all datasets and LLMs we tested, EAGLE-2 achieved the longest average acceptance length. Each drafting-verification cycle of EAGLE-2 generates approximately 4-5.5 tokens, significantly higher than other methods, roughly twice that of standard speculative sampling and Medusa. PLD and Lookahead have shorter average acceptance lengths, but since they either lack a draft model or their draft model is not a neural network, the overhead during the drafting phase is very low, resulting in a speedup ratio very close to their average acceptance length.

Medusa, Hydra, EAGLE, and EAGLE-2 have lower average acceptance lengths on QA (Natural Questions) and summarization (CNN/DM) tasks compared to other tasks, whereas standard speculative sampling does not show this reduction. The same pattern is observed for the speedup ratios. This discrepancy may be attributed to differences in the training data for the draft models. The draft model for standard speculative sampling uses both pretraining and SFT datasets, while Medusa, Hydra, EAGLE, and EAGLE-2 only use the SFT dataset. Natural Questions involves questions about world knowledge, such as ``Where was the 2015 rugby union world cup held?'', and world knowledge is primarily acquired through pretraining rather than SFT. Summarization tasks are also less represented in the SFT dataset. This suggests the potential benefits of expanding the draft model's training data. Despite this, EAGLE-2 still outperforms standard speculative sampling on these two datasets.

\begin{table*}[h]
  \centering
  \caption{Speedup ratios and average acceptance lengths $\tau$ of different methods. V represents Vicuna, L2 represents LLaMA2-Chat. SpS denotes standard speculative sampling, with its draft model being Vicuna-68M. Methods like Medusa relax acceptance conditions under non-greedy settings, which do not guarantee lossless acceleration. Therefore, we do not compare EAGLE-2 with these methods.}
  \resizebox{\linewidth}{!}{
    \begin{tabular}{cccccccccccccccc}
    \toprule
          &       & \multicolumn{2}{c}{MT-bench} & \multicolumn{2}{c}{HumanEval} & \multicolumn{2}{c}{GSM8K} & \multicolumn{2}{c}{Alpaca} & \multicolumn{2}{c}{CNN/DM} & \multicolumn{2}{c}{Natural Ques.} & \multicolumn{2}{c}{Mean} \\
    \midrule
    Model & Method & Speedup & $\tau$     & Speedup & $\tau$     & Speedup & $\tau$     & Speedup & $\tau$     & Speedup & $\tau$     & Speedup & $\tau$     & Speedup & $\tau$ \\
    \midrule
    \multicolumn{16}{c}{Temperature=0} \\
    \midrule
    \multirow{7}[2]{*}{V 13B} & SpS   & 1.93x & 2.27  & 2.23x & 2.57  & 1.77x & 2.01  & 1.76x & 2.03  & 1.93x & 2.33  & 1.66x & 1.88  & 1.88x & 2.18 \\
          & PLD   & 1.58x & 1.63  & 1.85x & 1.93  & 1.68x & 1.73  & 1.16x & 1.19  & 2.42x & 2.50  & 1.14x & 1.17  & 1.64x & 1.69 \\
          & Medusa & 2.07x & 2.59  & 2.50x & 2.78  & 2.23x & 2.64  & 2.08x & 2.45  & 1.71x & 2.09  & 1.81x & 2.10  & 2.07x & 2.44 \\
          & Lookahead & 1.65x & 1.69  & 1.71x & 1.75  & 1.81x & 1.90  & 1.46x & 1.51  & 1.46x & 1.50  & 1.36x & 1.39  & 1.58x & 1.62 \\
          & Hydra & 2.88x & 3.65  & 3.28x & 3.87  & 2.93x & 3.66  & 2.86x & 3.53  & 2.05x & 2.81  & 2.11x & 2.88  & 2.69x & 3.40 \\
          & EAGLE & 3.07x & 3.98  & 3.58x & 4.39  & 3.08x & 3.97  & 3.03x & 3.95  & 2.49x & 3.52  & 2.42x & 3.11  & 2.95x & 3.82 \\
          & EAGLE-2 & \textbf{4.26x} & \textbf{4.83} & \textbf{4.96x} & \textbf{5.41} & \textbf{4.22x} & \textbf{4.79} & \textbf{4.25x} & \textbf{4.89} & \textbf{3.40x} & \textbf{4.21} & \textbf{3.13x} & \textbf{3.74} & \textbf{4.04x} & \textbf{4.65} \\
    \midrule
    \multirow{4}[2]{*}{L2 13B} & PLD   & 1.42x & 1.46  & 1.63x & 1.70  & 1.41x & 1.44  & 1.16x & 1.20  & 1.42x & 1.45  & 1.12x & 1.15  & 1.36x & 1.40 \\
          & Lookahead & 1.58x & 1.64  & 1.80x & 1.85  & 1.65x & 1.69  & 1.47x & 1.50  & 1.46x & 1.53  & 1.42x & 1.45  & 1.56x & 1.61 \\
          & EAGLE & 3.03x & 3.90  & 3.76x & 4.52  & 3.20x & 4.03  & 3.01x & 3.83  & 2.70x & 3.59  & 2.83x & 3.47  & 3.09x & 3.89 \\
          & EAGLE-2 & \textbf{4.21x} & \textbf{4.75} & \textbf{5.00x} & \textbf{5.52} & \textbf{4.31x} & \textbf{4.90} & \textbf{4.13x} & \textbf{4.61} & \textbf{3.45x} & \textbf{4.24} & \textbf{3.51x} & \textbf{4.04} & \textbf{4.10x} & \textbf{4.68} \\
    \midrule
    \multirow{7}[2]{*}{V 7B} & SpS   & 1.82x & 2.36  & 1.99x & 2.61  & 1.71x & 2.26  & 1.65x & 2.21  & 1.81x & 2.44  & 1.60x & 2.16  & 1.76x & 2.34 \\
          & PLD   & 1.61x & 1.68  & 1.82x & 1.87  & 1.82x & 1.99  & 1.21x & 1.31  & 2.53x & 2.72  & 1.23x & 1.44  & 1.70x & 1.84 \\
          & Medusa & 1.91x & 2.52  & 2.02x & 2.67  & 1.89x & 2.59  & 1.79x & 2.48  & 1.42x & 2.02  & 1.51x & 2.09  & 1.76x & 2.40 \\
          & Lookahead & 1.63x & 1.69  & 1.72x & 1.77  & 1.84x & 1.99  & 1.38x & 1.57  & 1.44x & 1.53  & 1.45x & 1.60  & 1.58x & 1.69 \\
          & Hydra & 2.69x & 3.60  & 2.98x & 3.79  & 2.73x & 3.66  & 2.66x & 3.58  & 2.01x & 2.70  & 2.25x & 2.86  & 2.55x & 3.37 \\
          & EAGLE & 2.90x & 3.94  & 3.33x & 4.29  & 3.01x & 4.00  & 2.79x & 3.89  & 2.33x & 3.42  & 2.31x & 3.21  & 2.78x & 3.79 \\
          & EAGLE-2 & \textbf{3.62x} & \textbf{4.98} & \textbf{3.95x} & \textbf{5.33} & \textbf{3.63x} & \textbf{4.97} & \textbf{3.46x} & \textbf{4.86} & \textbf{2.94x} & \textbf{4.12} & \textbf{2.76x} & \textbf{3.82} & \textbf{3.39x} & \textbf{4.68} \\
    \midrule
    \multirow{4}[2]{*}{L2 7B} & PLD   & 1.38x & 1.43  & 1.52x & 1.59  & 1.32x & 1.37  & 1.15x & 1.19  & 1.48x & 1.52  & 1.15x & 1.20  & 1.33x & 1.38 \\
          & Lookahead & 1.61x & 1.66  & 1.72x & 1.77  & 1.58x & 1.65  & 1.49x & 1.52  & 1.49x & 1.54  & 1.48x & 1.53  & 1.56x & 1.61 \\
          & EAGLE & 2.78x & 3.62  & 3.17x & 4.24  & 2.91x & 3.82  & 2.78x & 3.71  & 2.43x & 3.41  & 2.61x & 3.44  & 2.78x & 3.71 \\
          & EAGLE-2 & \textbf{3.43x} & \textbf{4.70} & \textbf{4.03x} & \textbf{5.39} & \textbf{3.52x} & \textbf{4.77} & \textbf{3.45x} & \textbf{4.66} & \textbf{3.01x} & \textbf{4.12} & \textbf{3.15x} & \textbf{4.19} & \textbf{3.43x} & \textbf{4.64} \\
    \midrule
    \multicolumn{16}{c}{Temperature=1} \\
    \midrule
    \multirow{3}[2]{*}{V 13B} & SpS   & 1.62x & 1.84  & 1.72x & 1.97  & 1.46x & 1.73  & 1.52x & 1.78  & 1.66x & 1.89  & 1.43x & 1.70  & 1.55x & 1.82 \\
          & EAGLE & 2.32x & 3.20  & 2.65x & 3.63  & 2.57x & 3.60  & 2.45x & 3.57  & 2.23x & 3.26  & 2.14x & 3.06  & 2.39x & 3.39 \\
          & EAGLE-2 & \textbf{3.80x} & \textbf{4.40} & \textbf{4.22x} & \textbf{4.89} & \textbf{3.77x} & \textbf{4.41} & \textbf{3.78x} & \textbf{4.37} & \textbf{3.25x} & \textbf{3.97} & \textbf{3.07x} & \textbf{3.54} & \textbf{3.65x} & \textbf{4.26} \\
    \midrule
    \multirow{2}[2]{*}{L2 13B} & EAGLE & 2.68x & 3.45  & 2.89x & 3.78  & 2.82x & 3.67  & 2.66x & 3.55  & 2.41x & 3.39  & 2.37x & 3.31  & 2.64x & 3.53 \\
          & EAGLE-2 & \textbf{3.92x} & \textbf{4.51} & \textbf{4.58x} & \textbf{5.29} & \textbf{4.21x} & \textbf{4.80} & \textbf{3.85x} & \textbf{4.48} & \textbf{3.31x} & \textbf{4.08} & \textbf{3.43x} & \textbf{3.89} & \textbf{3.88x} & \textbf{4.51} \\
    \midrule
    \multirow{3}[2]{*}{ V 7B} & SpS   & 1.50x & 1.87  & 1.55x & 1.95  & 1.53x & 1.82  & 1.56x & 1.85  & 1.63x & 1.91  & 1.33x & 1.72  & 1.52x & 1.85 \\
          & EAGLE & 2.13x & 3.17  & 2.39x & 3.43  & 2.34x & 3.29  & 2.21x & 3.30  & 2.08x & 3.12  & 1.95x & 2.86  & 2.18x & 3.20 \\
          & EAGLE-2 & \textbf{3.05x} & \textbf{4.28} & \textbf{3.33x} & \textbf{4.65} & \textbf{3.07x} & \textbf{4.49} & \textbf{3.08x} & \textbf{4.43} & \textbf{2.63x} & \textbf{3.76} & \textbf{2.48x} & \textbf{3.56} & \textbf{2.94x} & \textbf{4.20} \\
    \midrule
    \multirow{2}[2]{*}{L2 7B} & EAGLE & 2.22x & 3.30  & 2.61x & 3.79  & 2.40x & 3.52  & 2.29x & 3.33  & 2.19x & 3.15  & 2.22x & 3.12  & 2.32x & 3.37 \\
          & EAGLE-2 & \textbf{3.19x} & \textbf{4.41} & \textbf{3.67x} & \textbf{5.06} & \textbf{3.35x} & \textbf{4.62} & \textbf{3.20x} & \textbf{4.48} & \textbf{2.73x} & \textbf{3.85} & \textbf{2.81x} & \textbf{4.01} & \textbf{3.15x} & \textbf{4.41} \\
    \bottomrule
    \end{tabular}%
    }
  \label{tab:big}%
\end{table*}%

\begin{table}[h]
  \centering
  \caption{Speedup ratios and average acceptance lengths $\tau$ with LLaMA2-Chat 70B, LLaMA3-Instruct 70B, and LLaMA3-Instruct 8B as the original LLMs, with the temperature set to 0, on the MT-bench dataset.}
  \resizebox{\linewidth}{!}{
    \begin{tabular}{cccc}
    \toprule
    Model & Method & Speedup & $\tau$  \\
    \midrule
    \multirow{4}[2]{*}{LLaMA2-Chat 70B} & PLD   & 1.31x & 1.39 \\
          & Lookahead & 1.52x & 1.64 \\
          & EAGLE & 3.01x & 3.81 \\
          & EAGLE-2 & \textbf{3.51x} & \textbf{4.48} \\
    \midrule
    \multirow{2}[2]{*}{LLaMA3-Instruct 70B} & EAGLE & 2.83x & 3.62 \\
          & EAGLE-2 & \textbf{3.29x} & \textbf{4.16} \\
    \midrule
    \multirow{2}[2]{*}{LLaMA3-Instruct 8B} & EAGLE & 2.72x & 3.65 \\
          & EAGLE-2 & \textbf{3.46x} & \textbf{4.53} \\
    \bottomrule
    \end{tabular}%
    }
  \label{tab:small}%
\end{table}%

\subsection{Ablation Study}

In this section, we conduct the ablation study.

\subsubsection{Value and Confidence Score}

EAGLE's draft model provides a good approximation of acceptance rates, but it is local and cannot reflect the actual probability of a draft token being accepted. Therefore, when selecting nodes for expansion, we use the value, which is the product of a draft token's confidence score and its ancestor nodes' confidence scores, as the basis for ranking. In this section, we compare the performance impact of expanding based on value versus confidence score. The experimental results in Table \ref{tab:aba} show that the speedup ratio and average acceptance length are both higher when expanding based on value, demonstrating the rationale behind the EAGLE-2 approach.

\subsubsection{Reranking}

The purpose of EAGLE-2's expansion phase is to deepen the draft tree, but the tokens selected may be globally less optimal than shallow nodes that were not selected. Therefore, during the reranking phase, we rerank all the draft tokens. We conducted an ablation study on this operation using the MT-bench and GSM8K dataset. As shown in Table \ref{tab:aba}, reranking improved both the average acceptance length and the speedup ratio.

\begin{table}[h]
  \centering
  \caption{Ablation experiment results with temperature set to 0 on Vicuna 7B. ``w/o value” indicates not using value and directly using confidence, ``w/o reranking” indicates not performing reranking, and ``w/o both” indicates neither value nor reranking is used.}
  \resizebox{\linewidth}{!}{
    \begin{tabular}{ccccc}
    \toprule
          & \multicolumn{2}{c}{MT-bench} & \multicolumn{2}{c}{GSM8K} \\
    \midrule
    Method & Speedup & $\tau$     & Speedup & $\tau$ \\
    \midrule
    w/o both & 2.81x & 3.92  & 2.85x & 3.93 \\
    w/o value & 3.21x & 4.39  & 2.93x & 3.96 \\
    w/o reranking & 3.48x & 4.86  & 3.50x & 4.85 \\
    EAGLE-2 & \textbf{3.62x} & \textbf{4.98} & \textbf{3.63x} & \textbf{4.97} \\
    \bottomrule
    \end{tabular}%
    }
  \label{tab:aba}%
\end{table}%

\section{Related Work}

With widespread applications of LLMs, there has been significant work \cite{liu2023deja} focused on accelerating LLM inference, such as low-bit quantization \cite{hubara2018quantized,shen2020q,kim2021bert,zadeh2020gobo,zafrir2019q8bert}, pruning \cite{gale2019state,sanh2020movement}, and knowledge distillation \cite{hinton2015distilling}. These methods reduce generation latency by decreasing the computational cost of each forward pass of the LLM. However, these approaches often degrade LLM performance to some extent, resulting in a trade-off between generation quality and computational overhead.

Speculative sampling methods achieve lossless acceleration by using the original LLM for verification. Early speculative decoding methods \cite{stern2018blockwise,sun2021instantaneous} accelerated generation in greedy settings, while \citet{leviathan2023fast,chen2023accelerating} proposed speculative sampling to extend the draft-verification framework to non-greedy generation. Subsequent work has largely focused on reducing draft overhead and enhancing consistency between the draft and the original LLM. SpecInfer \cite{miao2023specinfer} integrates multiple small models as the draft model, aggregating their drafts into a tree and using tree attention for parallel verification. Medusa \cite{cai2024medusa} trains a set of MLPs to parallelly predict multiple tokens using the original LLM's features, significantly reducing the latency during the drafting phase. EAGLE \cite{li2024eagle} autoregressively predicts feature sequences instead of token sequences and inputs the sampling results into the draft model to address uncertainty at the feature level, substantially improving the draft model's accuracy. This principle of eliminating uncertainty is also used in Hydra \cite{ankner2024hydra} and Recurrent Drafter \cite{zhang2024recurrent}. Parallel Decoding \cite{santilli2023accelerating}, Lookahead \cite{fu2023lookahead}, Ouroboros \cite{zhao2024ouroboros}, and CLLMs \cite{kou2024cllms} generate drafts using Jacobi iterations. Methods \cite{hooper2023speed,yang2023predictive,monea2023pass,li2024nearest,yi2024generation,liu2024kangaroo,sun2024triforce,elhoushi2024layer,svirschevski2024specexec} like Draft \& Verify \cite{zhang2023draft} utilize techniques such as layer skipping or early exit, using parts of the original LLM's parameters as the draft model. REST \cite{fu2024break} and LLMA \cite{yang2023inference} generate drafts through retrieval. Online Speculative Decoding \cite{liu2023online} and DistillSpec \cite{zhou2024distillspec} further align the draft model with the original LLM through additional training. Cascade Speculative Drafting \cite{chen2023cascade} and Staged Speculative Decoding \cite{spector2023accelerating} cascade draft models of different sizes.

Speculative sampling methods can achieve lossless acceleration, but they can also trade off quality for higher speedup ratios. For example, BiLD \cite{kim2024speculative} relaxes the acceptance conditions, while Medusa-2 \cite{cai2024medusa}, CLLMs \cite{kou2024cllms}, and SPACE \cite{yi2024generation} fine-tune the original LLMs.

Some works have already employed partially dynamic draft trees. BiLD \cite{kim2024speculative} and Kangaroo \cite{liu2024kangaroo} use early stopping based on the draft model's confidence to control the tree's depth. GLIDE and CAPE \cite{du2024glide} adds additional candidates when the top-1 token confidence is low, controlling the tree's depth, but the additional candidates are not further expanded, resulting in a structurally limited tree. In contrast, EAGLE-2 has no such limitations and can dynamically adjust the draft tree structure flexibly, leading to better performance.

\section{Conclusion}

In this paper, we introduce EAGLE-2, an efficient and \textbf{lossless} speculative sampling method. We found that EAGLE's draft model confidence is a good approximation of the acceptance rate for draft tokens. Based on this, EAGLE-2 employs a context-dependent draft tree structure, significantly increasing the number of accepted draft tokens and resulting in better speedup ratios. EAGLE-2 ensures that the generated results are consistent with the original LLMs and does not require additional training.  We conducted extensive evaluations using various LLMs across multiple datasets and compared EAGLE-2 with several state-of-the-art speculative sampling methods. In all our experiments, EAGLE-2 achieved the highest speedup ratios.

\bibliography{example_paper}
\bibliographystyle{icml2024}

\newpage
\appendix

\onecolumn

\section{Implementation Details}
\label{sec:id}
\textbf{Vanilla:} We use models from the Huggingface.transformers library with the PyTorch backend and pre-allocated KV cache. Other methods also use these models as their base.

\textbf{(Standard) Speculative Sampling:} We use the assisted generation feature from the HuggingFace Transformers library.

\textbf{PLD, Lookahead, Medusa, and Hydra:} We use the default settings and the officially released weights.

\textbf{EAGLE:} Vicuna and LLaMA2-Chat draft models use the officially released weights, while LLaMA3-Instruct is trained using the ShareGPT dataset (consistent with Medusa and Hydra).

\textbf{EAGLE-2:} For the 7B (8B), 13B, and 70B original LLMs, we set the total number of draft tokens to 60, 50, and 48, respectively, with a draft tree depth of 6, and select 10 nodes during the expansion phase.
\end{document}